\pdfoutput = 1
\documentclass[twocolumn]{article}

\usepackage{times}
\usepackage[numbers]{natbib}
\usepackage[english]{babel}
\usepackage{blindtext}
\usepackage{graphicx}
\usepackage{amsmath, amsthm, amssymb, bbm, bm}
\usepackage{enumerate}
\usepackage{float}      
\usepackage{subcaption}  
\usepackage{wrapfig}
\usepackage[margin=0cm]{caption}
\usepackage[titletoc, toc]{appendix}
\usepackage{tabularx}

\usepackage{multirow}
\usepackage{hhline}
\usepackage{makecell}
\usepackage{placeins}  

\usepackage[x11names, usenames, dvipsnames, svgnames, table]{xcolor}
\definecolor{firebrick}{rgb}{.698,.133,.133}
\definecolor{mybluelight}{rgb}{0.9, 0.9, 1.}
\definecolor{myorangelight}{rgb}{1., 0.9, 0.9}

\usepackage[utf8]{inputenc} 
\usepackage[T1]{fontenc}    
\usepackage{url}            
\usepackage{booktabs, colortbl}       
\usepackage{amsfonts}       
\usepackage{nicefrac}       
\usepackage{microtype}      

\usepackage{csquotes}
\usepackage{latexsym}

\usepackage{pifont}
\usepackage[boxruled, vlined, linesnumbered]{algorithm2e}
\SetAlFnt{\small}
\SetAlCapFnt{\small}
\SetAlCapNameFnt{\small}
\usepackage{algorithmic}
\algsetup{linenosize=\tiny}

\let\oldnl\nl
\newcommand{\nonl}{\renewcommand{\nl}{\let\nl\oldnl}}

\usepackage{paralist}

\usepackage{xspace}
\usepackage{soul}
\usepackage{dsfont}
\usepackage{stmaryrd}
\usepackage[textwidth=15mm]{todonotes}
\usepackage{dirtytalk}
\usepackage{pbox}
\usepackage{cprotect}

\usepackage{verbatim}
\usepackage{textcomp}
\usepackage[normalem]{ulem}

\usepackage{mathtools}
\usepackage{etextools}
\usepackage[inline]{enumitem}

\usepackage[colorlinks=true,allcolors=firebrick,bookmarks=false]{hyperref}

\definecolor{darkergreen}{RGB}{21, 152, 56}
\definecolor{red2}{RGB}{252, 54, 65}
\definecolor{Gray}{gray}{0.85}
\newcolumntype{g}{>{\columncolor{Gray}}c}

\let\OLDthebibliography\thebibliography
\renewcommand\thebibliography[1]{
  \OLDthebibliography{#1}
  \setlength{\parskip}{0pt}
  \setlength{\itemsep}{0pt plus 0.3ex}
}

\theoremstyle{definition}

\DeclarePairedDelimiterX{\divx}[2]{(}{)}{%
  #1\;\delimsize\|\;#2%
}

\usepackage{style}

\makeatletter
\@namedef{ver@everyshi.sty}{}
\newcommand{\removelatexerror}{\let\@latex@error\@gobble}

\title{Joint Multimodal Transformer for Emotion Recognition in the Wild}

\renewcommand\footnotemark{}

\author{
  Paul~Waligora$^{1*}$,
  ~Haseeb~Aslam$^{1*}$,
  ~Osama~Zeeshan$^{1}$,
  ~Soufiane~Belharbi$^{1}$,
  ~Alessandro~Lameiras~Koerich$^{1}$,
  \\~\textbf{Marco~Pedersoli}$^{1}$, 
  ~\textbf{Simon~Bacon}$^{2}$, and
  ~\textbf{Eric~Granger}$^{1}$\\
 	$^1$ LIVIA, Dept. of Systems Engineering, ETS Montreal, Canada \\
	$^2$ Dept. of Health, Kinesiology \& Applied Physiology, Concordia University, Montreal, Canada\\
{\footnotesize \textcolor{black}{*These authors contributed equally} } \\
{\tt\footnotesize \textcolor{black}{paul.waligora.1@ens.etsmtl.ca} }
}

\newcommand{\ignore}[1]{}

\begin{document}
\maketitle\thispagestyle{fancy}

\maketitle
\rhead{\color{gray} \small Waligora et al. \;  [CVPRw 2024]}

\begin{abstract}
Multimodal emotion recognition (MMER) systems typically outperform unimodal systems by leveraging the inter- and intra-modal relationships between, e.g., visual, textual, physiological, and auditory modalities. 
This paper proposes an MMER method that relies on a joint multimodal transformer (JMT) for fusion with key-based cross-attention. This framework can exploit the complementary nature of diverse modalities to improve predictive accuracy. Separate backbones capture intra-modal spatiotemporal dependencies within each modality over video sequences. Subsequently, our JMT fusion architecture integrates the individual modality embeddings, allowing the model to effectively capture inter- and intra-modal relationships.
Extensive experiments\footnote{Code: \href{https://github.com/PoloWlg/Joint-Multimodal-Transformer-6th-ABAW}{https://github.com/PoloWlg/Joint-Multimodal-Transformer-6th-ABAW}.} on two challenging expression recognition tasks -- (1) dimensional emotion recognition on the Affwild2 dataset (with face and voice) and (2) pain estimation on the Biovid dataset (with face and biosensors) -- indicate that our JMT fusion can provide a cost-effective solution for MMER. Empirical results show that MMER systems with our proposed fusion allow us to outperform relevant baseline and state-of-the-art methods. 
\end{abstract}


\begin{figure}[!t]
\centering
\includegraphics[scale=0.8]{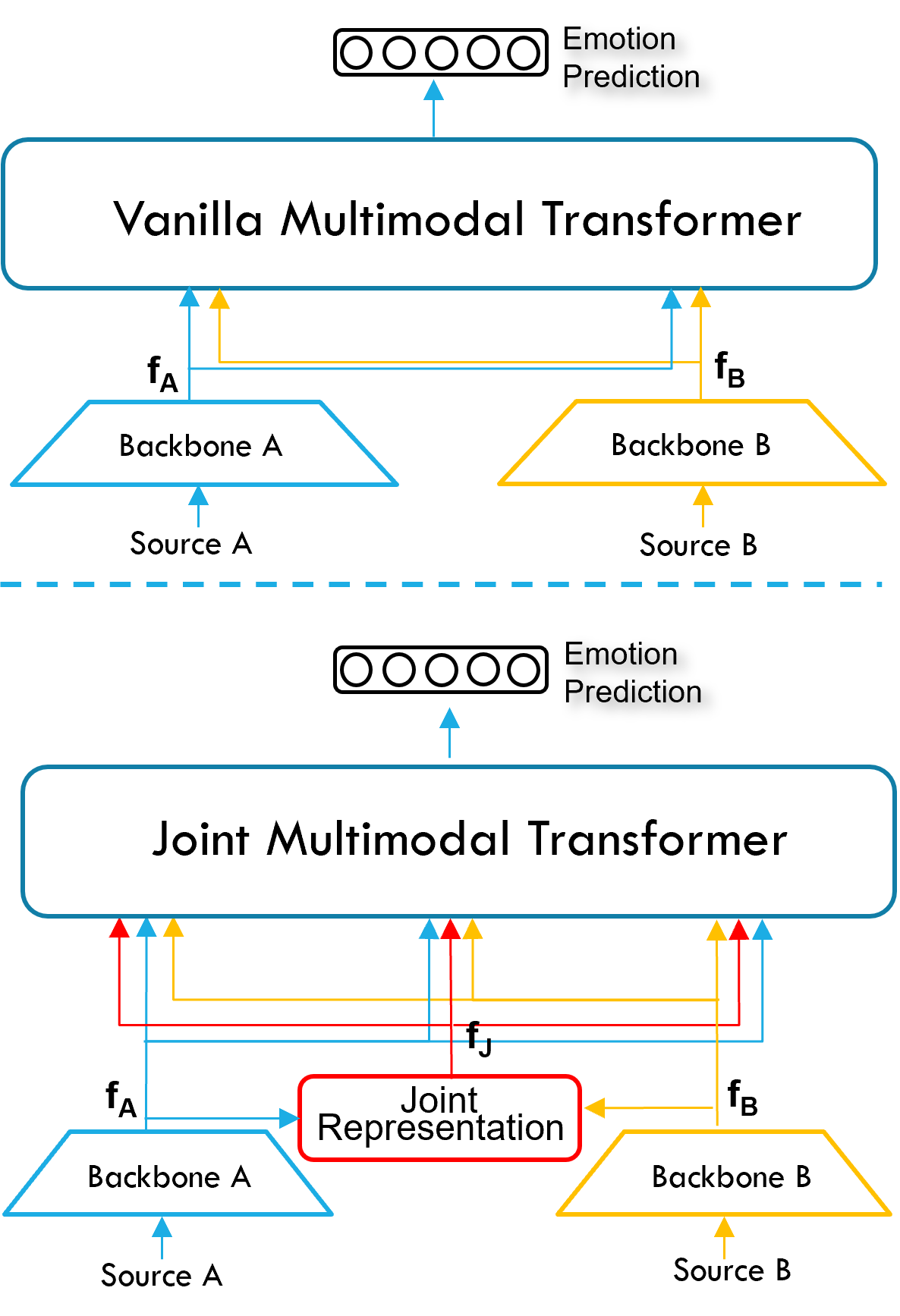}
  \caption{(Top) An illustration of the vanilla multimodal transformer fusion architecture in the case of two input sources, A and B. (Bottom) Our proposed JMT fusion (in red) relies on joint multimodal representations.}
  \label{fig:img/top_jmt.png}
\end{figure}

\section{Introduction}
\label{sec:intro}

Human-computer interaction is applied in a wide range of real-world scenarios, e.g., health care, the Internet of Things (IoT), and autonomous driving. Researchers have classified human emotions in different ways, most notably according to discrete categories, ordinal intensity levels, and the valence/arousal circumplex \cite{Anagnostopoulos}. With the recent advancements in deep learning and sensor technologies, research in affective computing has evolved from lab-controlled to real-world (in the wild) scenarios. In the latter, human emotions are usually expressed over a broader spectrum beyond the six basic categorical expressions - anger, disgust, fear, happy, sad, and surprise \cite{Ekman}. Therefore, there is much interest in analyzing and modeling complex and subtle expressions of emotions in real-world scenarios. For instance, the spectrum of emotions can be formulated as dimensional emotional recognition (ER), where complex human emotions are represented along arousal (intensity) and valence (positiveness) axes. 

Multimodal fusion has been widely explored for the problem of video-based ER in the literature \cite{cite6, 9320301}. For instance, audio and visual modalities may provide complementary and redundant information over a video sequence. These relationships must be captured to model the intricacies of human emotions effectively. Furthermore, effectively capturing both the intra-modal temporal dependencies within the audio and visual modalities and the inter-modal association across the audio and visual modalities is crucial in developing an effective AER system \cite{9856650}.

Several methods have been proposed for video-based ER and recurrent networks have been employed to capture the intra-modal temporal dependencies from video sequences \cite{cite6, 9320301, de2021mdn}. Recently, attention-based methods have been introduced to extract features that are the most relevant to downstream tasks. Cross-attention-based methods have also been \cite{9667055} employed to capture the inter-modal association between the audio, visual, and other modalities. Lu et al. \cite{lu2019vilbert} proposed ViLBERT, the seminal work in multimodal co-attention. Since then, many transformer-based cross-attention methods have been proposed \cite{Parthasarathy, wei-mult}. These methods, however, cannot effectively capture the intra-modal temporal dynamics. Further, they specialize in capturing the complementary information among the modalities but do not include a mechanism to explicitly capture the redundant information.

The proposed method introduces a third branch with the joint representation of the multiple modalities, as shown in Figure \ref{fig:img/top_jmt.png}. By incorporating a joint representation branch, the model can access additional contextual information that may not be fully captured by cross-attention alone. Such a joint representation branch can help improve the model's understanding of complex relationships between the input sequences. Further, the proposed method becomes more robust to noise or irrelevant information present in individual sequences, which helps mitigate the sensitivity of cross-attention to noisy inputs and improves the system's overall performance.

\noindent \textbf{Our main contributions are summarized as follows}.\\
\noindent\textbf{(1)} This paper proposes a joint multimodal transformer (JMT) fusion architecture that leverages joint modality representations. It captures inter- and intra-modal information in videos using key-based cross-attention, and exploits the redundant and complementary associations among modalities.
\noindent\textbf{(2)} An extensive set of experiments on two challenging emotion recognition datasets (pain estimation on BioVid and dimension valence-arousal assessment on Affwild2)  indicate that our proposed JMT fusion architecture can outperform relevant baseline and state-of-the-art methods.


\section{Related Work in Emotion Recognition}
\label{sec:rel_work}

\subsection{Multimodal Methods}
MMER refers to integrating multiple sources of information (modalities) to improve the accuracy and robustness of automated emotion recognition systems at the expense of complexity. These modalities typically include visual, audio, textual, and physiological. The seminal work in multimodal deep learning was proposed by Ngiam et al.~\cite{ngiam_mdl}, where the features from the audio and visual modalities were extracted separately, and then autoencoders and Restricted Boltzmann Machines were used to feature fusion. Tzirakis et al.~\cite{cite7} proposed one of the early approaches for A-V fusion for dimensional emotion recognition, in which the visual features were extracted using a ResNet50 and the audio features were obtained using a 1D convolutional neural network (CNN). The modality-specific features were concatenated and fed to a recurrent net for simultaneous temporal modeling and modality fusion. An empirical study was presented by Juan et al.~\cite{8914655}, where the authors studied the impact of fine-tuning multiple layers in a pretrained CNN for the visual modality. 

A two-stream autoencoder with a long short-term memory (LSTM) network was proposed by Nguyen et al.~\cite{9374787} to jointly learn and compact representative features from the visual and audio modalities. A knowledge distillation-based approach was investigated by Schonevald et al.~\cite{cite6} for visual modality. For the audio modality, spectrograms were obtained and fed to a CNN model, and the two modalities were fused using a recurrent net. A novel self-distillation scheme was put forward by Deng et al.~\cite{9607738} to overcome the problem of noisy labels in a multitasking setting. A two-stream aural visual (TSAV) network was proposed by Kuhnke et al.~\cite{9320301}, in which the audio features were extracted using a ResNet18, and the visual features were extracted using a 3D-CNN. The obtained embeddings were fed to a specially designed TSAV network for information fusion. 

Pain estimation is one of the primary problems in affective computing. Researchers have proposed many multimodal datasets for the pain estimation task. The facial activity descriptors method for pain estimation was introduced by Werner et al.~\cite{wener-multimodal-2014}. Dragomir et al.~\cite{dragomir-biovidA} propose a subject-independent method from facial images with a residual learning technique. A Sparse LSTM-based method was proposed by Zhi et al.~\cite{Zhi2021Multimodal_biovid} to solve the problem of vanishing gradients in temporal learning. Morabit et al.~\cite{morabit-biovida} proposed a data-efficient image transformer. To process multiscale electrodermal activity signals, a SE-Net-based network was proposed by Lu et al.~\cite{lu_biovid_phy}. Multimodal solutions to fuse the physiological and visual modalities were proposed by Werner et al.~\cite{wener-multimodal-2014}, Kachele et al.~\cite{Kchele2015BioVid_mm}, and Zhu et al.~\cite{Zhi2021Multimodal_biovid}. Physiological signals are more discriminative for pain classification than the visual modality.

\subsection{Attention-Based and Transformer Methods}
Since its inception, attention models have shown extraordinary performance in many applications. These models have been extensively investigated for capturing the inter and intra-modal associations between the audio and visual modalities for tasks like action localization \cite{lee-cross}, A-V event localization \cite{9423042}, and multimodal emotion recognition \cite{Parthasarathy}. An attention-based fusion mechanism was proposed by Zhang et al.~\cite{9320215}, 3D-CNNs and 2D-CNNs were used to extract multi-features in the visual modality, and for the audio modality, a 2D-CNN was used to learn representation from spectrograms. Specialized scoring functions were used to re-weight the audio and visual features. 

Recently, cross-modal attention has shown promising results because of its ability to model inter-modal relationships. Srinivas et al.~\cite{Parthasarathy} explored a transformer network with encoder layers, where cross-modal attention is used to fuse audio and visual features for continuous arousal/valence prediction in the wild. Tzirakis et al.~\cite{cite8} explored the idea of cross-attention in conjunction with self-attention. The authors proposed a transformer-based fusion architecture. Although the methods mentioned above have used cross-modal attention with transformers, they do not have any explicit mechanism to capture semantic relevance between the A-V features, particularly the intra-modal correlations. Zhang et al.~\cite{9607460} proposed a method for A-V fusion using leader-follower attentive fusion for continuous arousal/valence prediction. Attention weights are combined with the encoded visual and audio features. Cross attention presented in Praveen et al.~\cite{9667055} has shown a substantial increase in performance by using cross-correlation across the individual features. In contrast, our proposed method uses key-based cross-attention in multimodal transformers and explores the idea of feeding the joint A-V feature vector. By feeding the joint A-V feature representation, the proposed method effectively captures the inter- and intra-modal relationships simultaneously by interacting across itself and the other modalities. 

Huang et al.~\cite{huang_transformer} investigated the idea of multi-head attention in transformer-based fusion architecture, which was further combined with LSTM to capture the high-level representations. Tran et al.~\cite{tran_transformer} proposed a cross-modal transformer architecture that consisted of a multimodal cross-modal attention block, where the Queries were generated from one modality and the key values were generated from the other modality. Le et al.~\cite{le_transformer} put forward an end-to-end transformer-based fusion mechanism for multilabel multimodal emotion classification; the model consisted of three parts: i) three backbone networks for visual, audio, and textual feature extractor, ii) a transformer network for information fusion, and iii) classification network. Zhou et al.~\cite{zhou_transformer} proposed a transformer-based fusion scheme along with the temporal convolutional network (TCN); the audio and visual features were extracted using pretrained backbones followed by a TCN, the output of TCN was concatenated and fed to a transformer encoder block. A multilayer perceptron (MLP) was then used for the final prediction. 

All the aforementioned transformer-based fusion architectures primarily focus on intermodality correlation. In contrast, in addition to modeling the intermodality relationships to capture the complementarity between modalities, the proposed method explicitly feeds the joint (combined) features to the multimodal transformer to introduce redundancy. By incorporating this third joint representation branch, the proposed model can access enhanced contextual information that cross-attention might only partially capture. Doing this improves the model's understanding of complex relationships between the input sequences. Further, the proposed method becomes more robust to noise or irrelevant information present in individual sequences. This third joint representation allows the model to dynamically focus on this newly introduced information in sequences where both modalities are simultaneously noisy. This helps mitigate the sensitivity of cross-attention to noisy inputs and improves the system's overall performance.

\begin{figure*}
  \centering
  \includegraphics[scale=0.78]{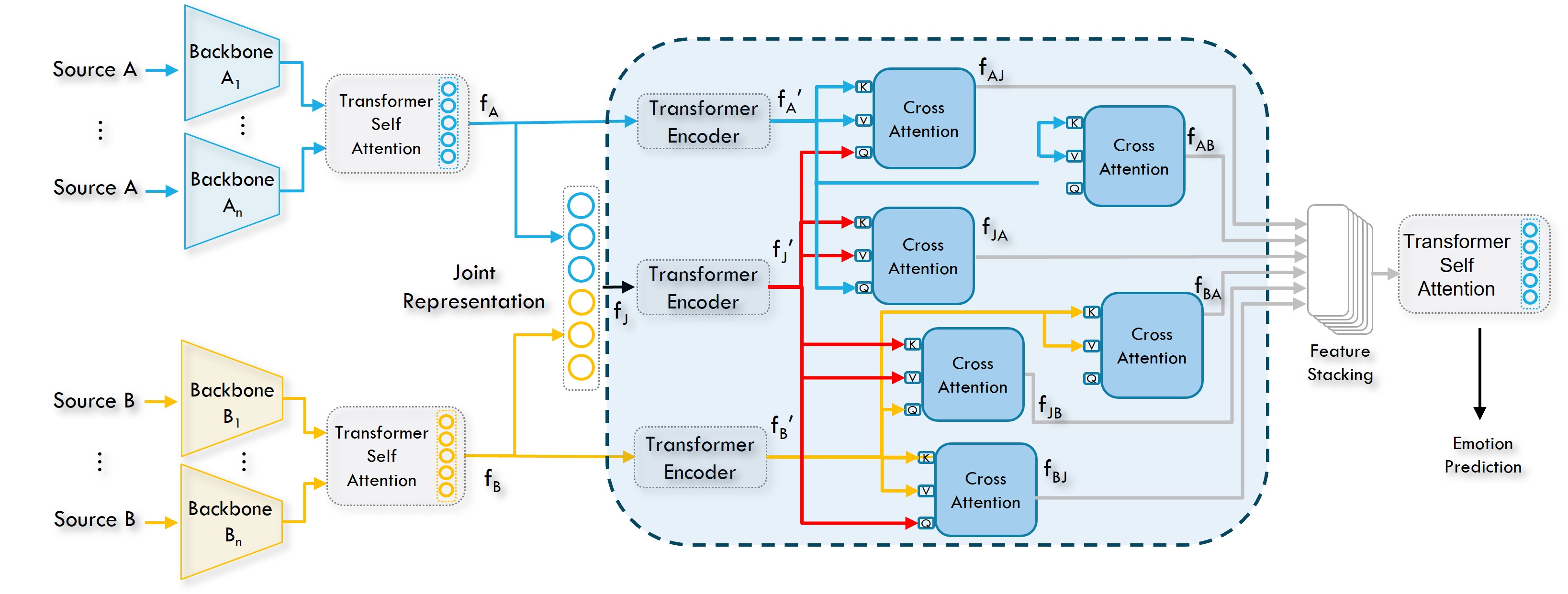}
  \caption{An overview of the proposed joint multimodal transformer model for A-V fusion. The audio and visual modalities are cross-attended using transformer blocks. The JMT block also takes in the joint representation (shown with red arrows). The output of the cross-attended features is concatenated, and an FC layer is used for valence/arousal prediction.}
  \label{fig:img/proposed.png}
\end{figure*}

\section{Proposed Approach}

The proposed method is a hierarchical fusion mechanism, where the intra-modality features are combined using transformer-based self-attention, and cross-modality features are fused using transformer-based cross-attention. Further, we feed a third joint representation to the joint transformer module (JTM) to enhance robustness. The $K$ (key matrix), $V$ (value matrix), and $Q$ (query matrix) vectors are shared among the six transformer blocks. In the end, the output of these six blocks is again fed to a transformer self-attention block to weigh the most relevant representations dynamically. The final prediction is made using fully connected (FC) layers.

\subsection{Modality Specific Feature Extraction}

In the first step, modality-specific features are extracted using backbones. The proposed method allows combining multiple backbones for each modality to improve system robustness. The extracted feature vectors from each backbone are fed to a transformer self-attention block. The combined feature vector represents the particular modality. 
For example, to capture information about a person's emotional state, we can use an R(2+1)D CNN pretrained on the Kinetics-400 dataset \cite{kuhnke2020two} to extract visual features. For the audio modality, we could extract features using a ResNet18 \cite{he2016deep} CNN with a GRU \cite{cho2014learning}.

\noindent \subsection{Multi-transformer Attention Fusion}

We define $\boldsymbol{f_A}$ as the deep features extracted from backbone A, and $\boldsymbol{f_B}$ as the deep features extracted from the backbone B in Figure \ref{fig:img/proposed.png}. Given $\boldsymbol{f_B}$ and $\boldsymbol{f_A}$, the joint feature representation is obtained by concatenating $\boldsymbol{f_B}$ and $\boldsymbol{f_A}$ feature vectors:  
\begin{equation}
 {\boldsymbol f_J} = [{\boldsymbol f}_{\mathbf B} ; {\boldsymbol f}_{\mathbf A}]\;,
\end{equation}

\noindent where ${[\cdot ; \cdot]}$ denotes a concatenation operation.
The concatenated features are then fed to an FC layer for dimensionality reduction of the joint feature representation to yield $\boldsymbol{f_{J}}$. We now have three key sources of information: ${\boldsymbol f}_{\mathbf B}$, ${\boldsymbol f}_{\mathbf A}$ and $\boldsymbol{f_{J}}$, which have the same dimensionality.

Each representation is then fed to a specific encoder. Our model is composed of three different encoders, one for each type of feature ${\boldsymbol f}_{\mathbf B}$, ${\boldsymbol f}_{\mathbf A}$, and $\boldsymbol{f_{J}}$.
Each encoder consists of a multi-head-self-attention (Eq. \ref{eq:mh_equation}) followed by a fully connected feed-forward network. Residual connection and layer normalization are performed around both of these layers. 
The key is used to associate a sequence to a key value, the value matrix holds information that is ultimately used to compute the output of the attention mechanism, and the query matrix represents a set of vectors used to query the key-value pairs. The ${K}$, ${V}$, ${Q}$ matrix are calculated this way: ${K = {X}{W_{K}}}$, ${Q = {X}{W_{Q}}}$, ${V = {X}{W_{V}}}$. ${X}$ corresponds to one of the sources of information and ${W_{K}}$, ${W_{Q}}$, ${W_{V}}$ are the weights of the key, query, value matrices respectively.
The output values of the self-attention layers are given by:
 \begin{equation}
     {\mbox{Attention}(Q,K,V)= \mbox{softmax}({K}{Q^T}/{d_{k}})V}
     \label{eq:mh_equation}
 \end{equation}

With self-attention layers, each encoder focuses independently on important cues related to its respective source of information. Afterward, each encoder embedding is combined by utilizing six cross-modal attention layers where the query matrix Q is shared with the key K and value V matrix of the other source of information. Sharing this matrix between each source of information helps the model add redundancy and complement the visual and audio modalities, thus improving its performance. At the output of each of the six cross-attention modules, the feature vector of dimension 512 is output. These six feature vectors are then stacked to form a sequence, which is then fed to a transformer self-attention block. This block dynamically selects and weighs these feature vectors. The final attended features are fed to an FC layer for final prediction.

The model aims to maximize the Concordance Correlation Coefficient (CCC) \cite{lawrence1989concordance}, which is common in dimensional emotion recognition. To achieve this, we minimize the following loss:
\begin{equation}
 \\{L}_c = 1 - \rho_c = \ 1 - \frac{2\rho_{xy}^2}{\rho_x^2 + \rho_y^2 + (\mu_x - \mu_y)^2} \
\end{equation}
\noindent where ${\rho_{xy}^2}$ is the covariance between the predictions and the ground truth, ${\rho_x^2}$ and ${\rho_y^2}$ the variance of the prediction and the ground truth, ${\mu_x}$ and ${\mu_y}$ the mean of prediction and the ground truth.

\section{Experimental Methodology}

\subsection{Datasets}

\paragraph{Affwild2} was put forward by \cite{Kollias}. It is one of the largest and most comprehensive datasets for affective computing focused in-the-wild scenarios. The dataset is also a part of the Affective Behavior Analysis in the Wild (ABAW) challenge \cite{aslam2024distilling,kollias20246th, kollias2023abaw2, kollias2023multi, kollias2023abaw, aslam2023privileged, kollias2022abaw, kollias2021analysing, kollias2021affect, kollias2021distribution, kollias2020analysing, kollias2019expression, kollias2019deep, kollias2019face, zafeiriou2017aff}. The dataset comprises $564$ videos that were collected from YouTube. These videos are labeled for three main affective computing tasks: i) categorical expression recognition, ii) continuous valence arousal prediction, and iii) action unit detection. The dataset comes with a train, validation, and test split with 341, 71, and 152 videos, respectively. The continuous valence/arousal annotation set is used to validate the proposed method.   

\paragraph{Biovid Heat Pain Database:} 
 The BioVid Heat Pain Database \cite{werner2014automatic} comprises 87 subjects where heat pain was induced experimentally at the right arm with four different intensities. The data comprises video and depth map video from a Kinect camera, galvanic skin response (EDA), electromyograph (EMG) on trapezius muscle, and electrocardiogram (ECG). The Biovid dataset has various partitions from Part A through E. These partitions differ in the modalities, annotations, and tasks. We use Part A of the dataset and utilize the video and raw EDA data to validate our proposed method.

\subsection{Implementation Details}

\paragraph{Affwild2:}
In the visual modality, cropped and aligned facial images provided with the dataset are used \cite{kollias2021analysing}.  Black frames (zero pixels) replace missing frames in the visual modality. These images are then fed to a 3D network that takes the input size of 224$\times$224.
A clip length of 8 is used, which makes up a sub-sequence of 64 frames. Each sub-sequence contains eight clips. A dropout with a value of 0.8 was used in the linear layers for network regularization.  The network was optimized using the stochastic gradient descent (SGD) iterative method with an initial learning rate (LR) of $10^{-3}$. The batch size used for the visual modality was 8. For further generalization, random cropping and random horizontal flips were added as data augmentation. The maximum number of epochs is 50, with early stopping for model selection. In the audio modality, the audio from each video is separated and resampled for 44 kHz. Following the segmentation in the visual modality, small vocal segments are extracted corresponding to the length of the clip. 
A ResNet18 is used to extract the features from the spectrograms. A Discrete Fourier transform with a length of 1024, a hop length of 10 msec, and a window length of 20 msec is used to obtain spectrograms of each vocal segment. The spectrograms have a resolution of 64$\times$107 pixels, corresponding to a single clip in the visual modality. Other preprocessing of spectrograms include mean and variance normalization, as well as conversion to log-power spectrum.
The first convolutional layer in the pretrained ResNet model is adapted to take in the single-channel spectrograms. The learning rate of $1\times10^{-2}$ is used and Adam optimizer is used to optimize the network. 64 batch size is set for the audio modality.

The audio and visual backbones are frozen to train the A-V fusion network and only train the whole transformer-based fusion model. Each of these backbones outputs deep feature vectors of dimension 512. These features are concatenated (to obtain the joint feature representation, a feature vector of dimension 1024) and then fed to an FC layer to reduce dimensionality to 512. Each of these features is fed to the fusion model, as we can see in Figure \ref{fig:img/proposed.png}. At the output of each of the six cross-attention modules, the feature vector of dimension 512 is outputted. These six feature vectors are then stacked to form a sequence, which is then fed to a transformer self-attention block. This block dynamically selects and weighs these feature vectors. The final attended features are fed to an FC layer for final prediction.
We perform a grid search to find the optimal learning rate and batch size for our fusion network. Thus, we use a list of learning rates: [$8\times10^{-4}$, $6\times10^{-4}$,   $3\times10^{-4}$] and an SGD optimizer to optimize the fusion network. We use a batch size of 32, and the maximum number of epochs is 5, with early stopping for fusion model selection. We select the best model among the learning rates listed before. 

\paragraph{Biovid:} 
In the visual modality, the faces are cropped and aligned using an MTCNN. We apply the frame retention strategy for missing frames where the face is not visible, and the MTCNN cannot capture any frame. Further, to ensure noise-free input to the visual model, we clip the first 2 seconds of the video and the last 0.5 seconds at the end because the subjects show no signs of pain during this duration. The total number of frames is 75. The extracted faces are fed to an R3D model for a visual feature extractor. The batch size is set to 64. The network is separately optimized using the SGD optimizer and the learning rate of $10^{-3}$.

\begin{table}[b!]
\renewcommand{\arraystretch}{1.2}
  \centering
  \caption{Description of the architecture of the custom 1D-CNN for the physiological backbone.}
  \begin{tabular}{|c|c| c |c |c|}
    \hline
    \textbf{Layer} & \textbf{No. of} & \textbf{Size of} &  &  \\
    \textbf{type} & \textbf{Filters}  & \textbf{Kernel} & \textbf{Stride} & \textbf{Output}  \\
     \hline\hline
     \makecell{Input} & - & - & - &   $2816$ $\times$ $1$ \\
    \hline
     1st Conv & $32$ &  $5$ & $2$ & $1406$ $\times$ $32$ \\ 
     ReLU & - & -  & - & $1406$ $\times$ $32$ \\ 
     Max Pooling & - & $2$  & - & $703$ $\times$ $32$\\ 
    \hline
     2nd Conv & $64$ &  $5$ & $1$ & $699$ $\times$ $64$ \\ 
     ReLU  & - & -  & - & $699$ $\times$ $64$ \\ 
     Max Pooling & - & $2$  & - & $349$ $\times$ $64$\\ 
    \hline
     1st FC & - &  - & - & $512$ \\ 
     ReLU  & - & -  & - & $512$ \\ 
    \hline
     2nd FC  & - &  - & - & $2$ \\ 
    \hline
  \end{tabular}
  \label{table:table-1dcnn}
\end{table}

For the physiological modality, the EDA is used. The signal is clipped to correspond to the visual modality and fed to a 1D CNN. The architecture of the custom 1D-CNN is shown in Table \ref{table:table-1dcnn}. The CNN outputs a 512-dimensional feature vector. The model is optimized using the SGD optimizer with a learning rate of $10^{-4}$. The batch size for the physiological backbone is set to 1024.

\begin{figure}[b!]
\centering
\includegraphics[scale=0.89]{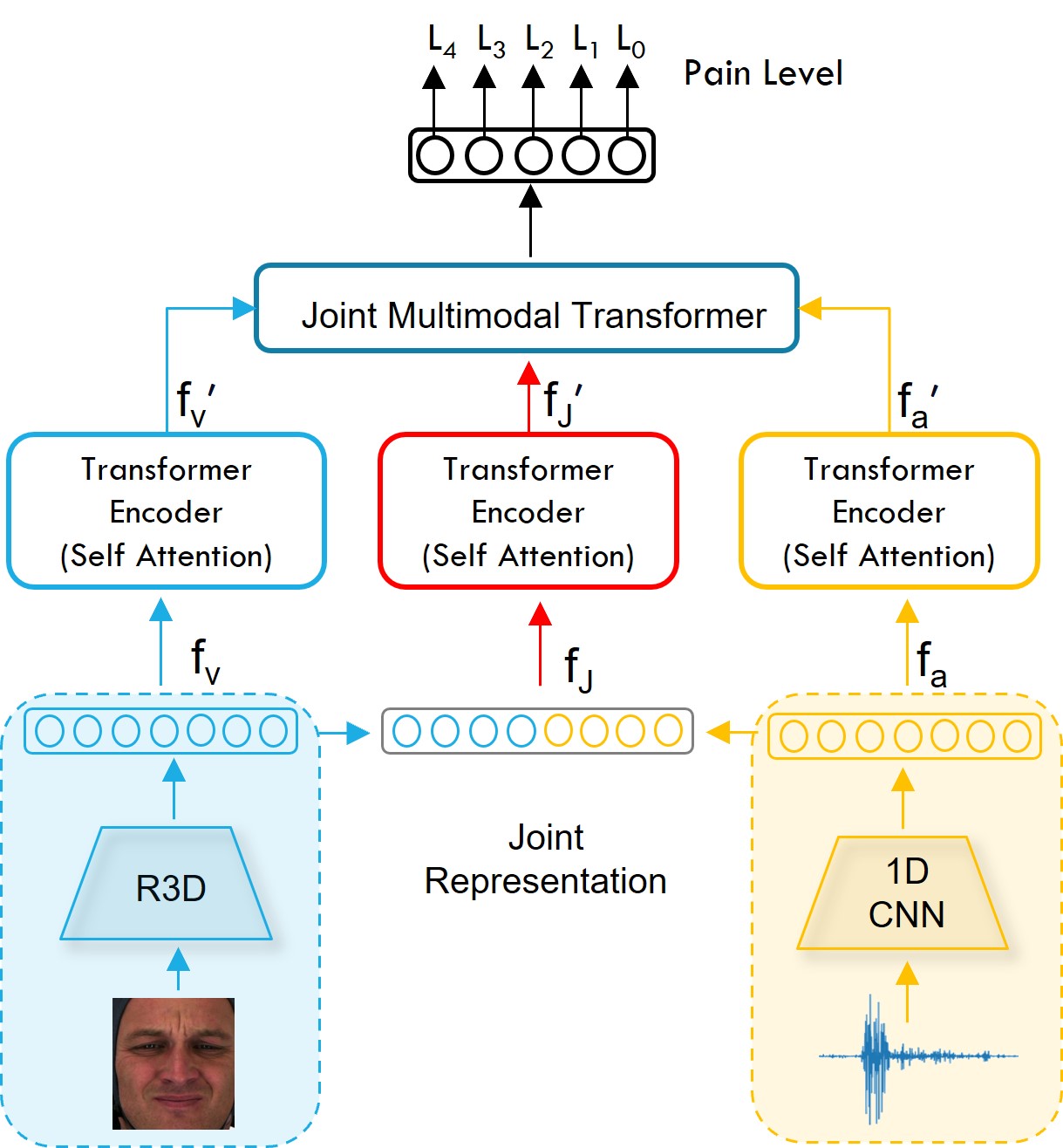}
  \caption{Illustration of the proposed joint multimodal transformer architecture used for the Biovid pain estimation task. The blue branch shows the visual backbone, and the yellow branch is the physiological backbone. The joint representation is shown with a red block. The three feature vectors are fed into the joint transformer block.}
  \label{fig:img/bio-img.png}
\end{figure}

For the modality fusion, the two feature vectors and the joint representation are fed to the joint transformer block, as shown in Figure \ref{fig:img/bio-img.png}. The FC layers are removed from both backbones that were added in the backbone training phase. 512-dimensional feature vectors from visual and physiological backbones are obtained and fed to the joint multimodal transformer module. The backbones are frozen, and the joint transformer block is optimized using the ADAM optimizer with a learning rate of $5\times10^{-6}$, and the batch size is set to 128.

\section{Results and Discussion}

\subsection{Comparison with the State-of-the-Art}

This section compares the proposed method with the baseline and state-of-the-art on the Affwild2 and Biovid datasets.

\begin{table*}[htpb!]
\renewcommand{\arraystretch}{1.}
  \centering
  \caption{A comparison of the proposed JMT and state-of-the-art fusion models for pain estimation on the Biovid Heat Pain Database. The highest score is indicated in bold. }
  \begin{tabular}{|p{6cm}|c|c|c|}
    \hline
    \multicolumn{1}{|c|}{\textbf{Method}} & \textbf{Modality} & \textbf{XV Scheme}  & \textbf{Accuracy (\%)} \\
    \hline\hline
    Werner et al.~\cite{werner2014automatic} \it{ICPR 2014} & EDA, ECG, EMG + Video & 5-FOLD & 80.6\\
    \hline
    Werner et al.~\cite{werner2016automatic} \it{IEEE TAC 2016} & Video  & LOSO & 72.4\\
    \hline
    Kachele et al.~\cite{kachele2016methods} \it{IEEE IJSTSP 2016} & EDA, ECG, EMG & LOSO & 82.73\\
    \hline
    Lopez et al.~\cite{lopez2017multi} \it{ACIIW 2017} & EDA, ECG & 10-FOLD & 82.75\\
    \hline
    Lopez et al.~\cite{lopez2018continuous} \it{EMBC 2018} & EDA & LOSO & 74.21\\
    \hline
    Thiam et al.~\cite{thiam2019exploring} \it{Sensors 2019} & EDA & LOSO & 84.57\\
    \hline
    Wang et al.~\cite{wang2020hybrid} \it{EMBC 2020} & EDA, ECG, EMG & LOSO & 83.3\\
    \hline
    Pouromran et al.~\cite{pouromran2021exploration} \it{PLoS ONE 2021} & EDA & LOSO & 83.3\\
    \hline
    Thiam et al.~\cite{thiam2021multi} \it{Frontiers 2021} & EDA, ECG, EMG & LOSO & 84.25\\
    \hline
    Phan et al.~\cite{phan2023pain} \it{IEEE Access 2023} & EDA, ECG, EMG & LOSO & 84.8\\
    \hline
    \hline
    Audio backbone: 1D CNN only & EDA & 5-FOLD & 77.2\\
    \hline
    Visual backbone: R3D model only & Video & 5-FOLD & 72.9\\
    \hline
    Fusion: feature concatenation & EDA + Video & 5-FOLD & 83.5\\
    \hline
    Fusion: vanilla transformer & EDA + Video  & 5-FOLD & 87.8\\
    \hline
    Fusion: JMT (ours) & EDA + Video & 5-FOLD  & \textbf{89.1}\\
    \hline    
  \end{tabular}
  \label{table:table_biovid}
\end{table*}

Table \ref{table:table_biovid} presents the performance of our fusion model on the Biovid Heat Pain Database. We performed 5-fold cross-validation to pick up the best average fusion model. It can be seen from the table that the proposed model can achieve state-of-the-art performance with multimodal input while using 5-fold cross-validation. We used 5-fold cross-validation instead of LOSO validation method due to the computational cost of processing video. The physiological modality is stronger in the Biovid database. Many studies have validated the models on physiological modality. Our empirical results show that the EDA-only accuracy is 77.2\%, whereas the visual-only accuracy is 72.9\%. The proposed model can improve over unimodal performance and achieves state-of-the-art performance on the Biovid dataset. We also compare it with standard fusion techniques like feature concatenation. For a fair comparison, we keep all the parameters the same. The proposed model improves 6\% over simple feature concatenation and 1.3\% over a vanilla multimodal transformer i.e. without a joint representation. 

\begin{figure}[b!]
\centering
\includegraphics[scale=0.635]{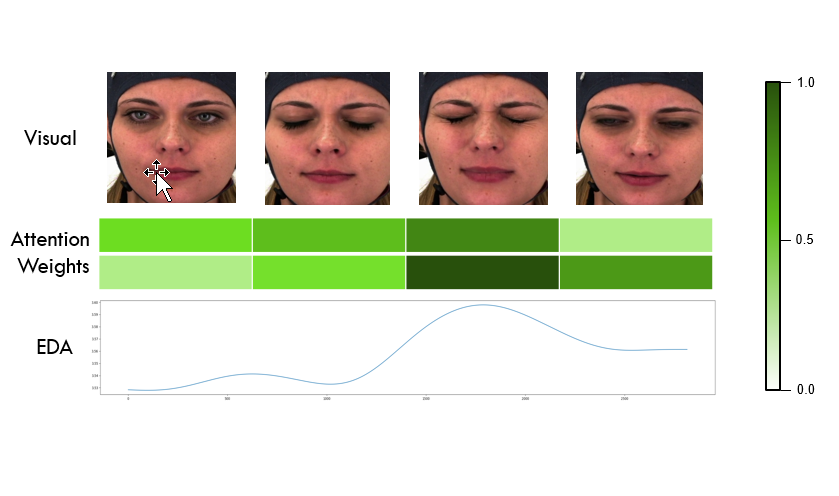}
  \caption{Visualization of attention weights for visual and physiological modalities on the Biovid heat pain database. The facial frames are taken 1400 msec each.}
  \label{fig:img/awv.png}
\end{figure}

Figure \ref{fig:img/awv.png} shows the visualization of the attention weights generated by the joint transformer model. It can be seen that the model gives more weightage to the physiological modality.

Table \ref{table:table_aff} shows the valence and arousal CCC values on the Affwild2 official validation set and custom-defined folds to increase generalizability. On the official validation set, the proposed method achieves a 0.666 average with 0.717 valence and 0.614 arousal. In addition, we performed ensembling by considering a dual modeling. We observed that different training configurations lead to models that can be best at once case: 'Valence', or 'Arousal'. Therefore, we used a dual model for prediction by taking the best at each category. This lead to performance increase.

\begin{table}[htpb!]
\renewcommand{\arraystretch}{1.4}
  \centering
  \caption{CCC for valence and arousal of the fusion model trained on different folds of the Affwild2 validation set. Highest scores are indicated in bold.}
  \begin{tabular}{|c|c|c|c|}
    \hline
    \textbf{Validation Set } & \textbf{Valence} & \textbf{Arousal}  & \textbf{Average} \\
    \hline\hline
    Official & 0.717 & 0.614 & 0.666\\
    \hline
    fold-1 & 0.705  & 0.683 & 0.694\\
    \hline
    fold-2 & 0.741 & 0.623 & 0.682\\
    \hline
    fold-3 & 0.657 & 0.637 & 0.647\\
    \hline
    fold-4 & 0.760 & 0.666 & 0.713\\
    \hline
    fold-5 & 0.684 & 0.629 & 0.657\\
    \hline
    Ensembling & \textbf{0.769} & \textbf{0.692} & \textbf{0.731}\\
    \hline
  \end{tabular}
  \label{table:table_aff}
\end{table}

Table \ref{table:table_affwild_test} compares the proposed JMT fusion method against state-of-the-art fusion methods on the Affwild2 test set. The proposed method can achieve 0.472 valence and 0.443 arousal. The average of valence and arousal is 0.458. The proposed method significantly improves over the baseline (provided by the challenge organizers). It is essential to mention here that the other methods that achieve higher performance are due to extensive pertaining, the use of additional modalities like text, and the use of more robust backbones. For a fair comparison, we use a similar setting to Joint Cross Attention~\cite{praveen2022joint}, which includes similar pretraining and identical backbones for audio and visual modalities. They used joint cross-attention to fuse the two modalities and achieve a 0.369 average. On the other hand, the proposed model uses JMT fusion and can achieve a 0.443 average. The proposed method improves 7\% over the joint cross-attention-based method. 

\begin{table}[htpb!]
\renewcommand{\arraystretch}{1.5}
  \centering
  \caption{CCC performance of the proposed JMT and state-of-the-art methods for A-V fusion on the Affwild2 test set.}
  \begin{tabular}{|l|c|c|c|}
    \hline
    \multicolumn{1}{|c|}{\textbf{Method}}  & \textbf{Valence}  & \textbf{Arousal} & \textbf{Mean} \\
    \hline\hline
    Baseline   & 0.180  & 0.170 & 0.175\\

    \hline
    \makecell[l]{Joint Cross \\ Attention \cite{praveen2022joint}}    & 0.374  & 0.363 & 0.369\\
    
    \hline
    AU-NO \cite{karas2022continuous}  & 0.418  & 0.407 & 0.413\\
  
    \hline
       HSE-NN \cite{savchenko2022frame}  & 0.417  & 0.454 & 0.436\\
 
    \hline

      PRL \cite{nguyen2022ensemble} & 0.450  & 0.445 & 0.448\\
      \hline
             \makecell[l]{Joint Multimodal \\ Transformer (ours)}   & 0.472  & 0.443 & 0.458\\
    \hline
    FlyingPigs \cite{zhang2022continuous}    & 0.520  & \textbf{0.602} & 0.561\\
    \hline
        Situ-RUCAIM3 \cite{meng2022multi}   & \textbf{0.606}  & 0.596 & \textbf{0.601}\\
    \hline
  \end{tabular}
  \label{table:table_affwild_test}
\end{table}

\subsection{Ablations}

Table \ref{table:table_affwild_test2} shows that the proposed model improves performance over the vanilla multimodal transformer by 1.8\% with an R2D1 and ResNet18 backbone on the Affwild2 dataset. With an I3D and a ResNet18 backbone, results are improved by 1.7\%, as shown in Table \ref{table:table_affwild_test2_I3D}. Notice that the results on the test set are different from Table \ref{table:table_affwild_test} because we used a different train/validation split. Additionally, results show that feature-level fusion of 2 vision backbones (I3D and R2D1) improves the performance compared to using a single vision backbone.

\begin{table}[t!]
\renewcommand{\arraystretch}{1.}
  \centering
  \caption{CCC performance of the backbones alone, and combined using baselines and the proposed A-V fusion method on the Affwild2 test set. Experiments are performed on the default training-validation split. R2D1 (visual) and ResNet18 (audio) backbones are used for all cases.}
  \begin{tabular}{|l|c|c|c|}
    \hline
    \multicolumn{1}{|c|}{\textbf{Fusion Model}}  & \textbf{Valence}  & \textbf{Arousal} & \textbf{Mean} \\
    \hline\hline
     \makecell[l]{R2D1 model only \\ (visual)}   & 0.194  & 0.310 & 0.252\\
     \hline
     \makecell[l]{ResNet18 model only \\ (audio)}   & 0.273  & 0.246 & 0.260\\
     \hline
     \makecell[l]{Concat + FC layers }   & 0.320  & 0.327 & 0.323\\
     \hline
     \makecell[l]{Vanilla Transformer}   & \textbf{0.376}  & 0.334 & 0.355\\
     \hline
     \makecell[l]{JMT (ours)}   & 0.366  & \textbf{0.379} & \textbf{0.373}\\
     
    \hline
  \end{tabular}
  \label{table:table_affwild_test2}
\end{table}

\begin{table}[t!]
\renewcommand{\arraystretch}{1.}
  \centering
  \caption{CCC performance of visual and audio backbones alone, and their A-V fusion using the proposed and baseline methods on the Affwild2 test set. Experiments are performed on the default training-validation split. I3D (visual) and ResNet18 (audio) backbones are used for all cases. FC: fully connected layers for feature-level fusion of backbones from the same modality. TR: transformer for fusing features from backbones of the same modality.}
  \begin{tabular}{|l|c|c|c|}
    \hline
    \multicolumn{1}{|c|}{\textbf{Fusion Model}}  & \textbf{Valence}  & \textbf{Arousal} & \textbf{Mean} \\
    \hline\hline
    \makecell[l]{Joint Cross Attention \cite{praveen2022joint}} & 0.374  & 0.363 & 0.369\\
    \hline
     \makecell[l]{I3D model only \\ (visual)}   & 0.336  & 0.422 & 0.379\\
     \hline
     \makecell[l]{ResNet18 model only \\ (audio)}   & 0.273  & 0.246 & 0.260\\
     \hline
     \makecell[l]{Concatenation + FC}   & 0.387  & 0.453 & 0.420\\
     \hline
     \makecell[l]{Vanilla Transformer  \\
     - I3D (visual) \\ 
     - ResNet18 (audio)}   & 0.432  & 0.410 & 0.421\\
     \hline
     \makecell[l]{JMT (ours) \\
     - I3D (visual) \\ 
     - ResNet18 (audio)}   & 0.425  & 0.450 & 0.438\\
     \hline
     \makecell[l]{JMT (ours) \\ 
     - I3D+R2D1 (visual, FC) \\
     - ResNet18 (audio)}   & \textbf{0.472}  & 0.443 & \textbf{0.458}\\
    \hline
    \makecell[l]{JMT (ours) \\ 
     - I3D+R2D1 (visual, TR) \\
     - ResNet18 (audio)}   & 0.458  & \textbf{0.445} & 0.452\\
    \hline
  \end{tabular}
  \label{table:table_affwild_test2_I3D}
\end{table}

Table \ref{tab:ablation} shows the results of the proposed method with and without the joint representation. On the Biovid dataset, the joint multimodal transformer improves by 1.3\% over the vanilla multimodal transformer. 

\begin{table}[htpb!]
\renewcommand{\arraystretch}{1.5}
\centering
    \caption{Performance of the proposed approach compared to the vanilla multimodal transformer on test set. Valence and arousal for the Affwild2 dataset and accuracy for the Biovid dataset. I3D + ResNet18 backbones are used for the Affwild2 dataset and R3D + 1D CNN are used for the BioVid dataset. Default training/validation split is used in the Affwild2 dataset and 5-fold cross validation is performed on the BioVid dataset.}
        \begin{tabular}{|c|c|c|}
            \hline
            \multirow{1}{*}{\textbf{Database}} & \multicolumn{1}{|c|}{\textbf{Method}} & \multicolumn{1}{c|}{\textbf{Accuracy}} \\\hline\hline
            \multirow{3}{*}{\textbf{Affwild2}} & \multicolumn{1}{|l|}{\makecell[l]{Vanilla Multimodal \\ Transformer}}  & \multicolumn{1}{l|}{\makecell[r]{V: \textbf{0.432} \\ A: 0.410 \\ Avg: 0.421}}\\\cline{2-3}
                                 & \multicolumn{1}{|l|}{\makecell[l]{Joint Multimodal \\ Transformer}} &  \multicolumn{1}{l|}{\makecell[r]{V: 0.425 \\ A: \textbf{0.450} \\ Avg: \textbf{0.438} }}  \\\hline
            \multirow{3}{*}{\textbf{Biovid}} & \multicolumn{1}{|l|}{\makecell[l]{Vanilla Multimodal \\ Transformer}} & \multicolumn{1}{c|}{87.8} \\\cline{2-3}
                                 & \multicolumn{1}{|l|}{\makecell[l]{Joint Multimodal \\ Transformer}} & \multicolumn{1}{c|}{\textbf{89.1}}  \\\hline
        \end{tabular}
    \label{tab:ablation}
\end{table}

\section{Conclusion}
Multimodal emotion recognition systems outperform their unimodal counterparts, especially in the wild environment. The missing and noisy modality is a prevalent issue with in-the-wild emotion recognition systems. Many attention-based methods have been proposed in the literature to overcome this problem. These methods aim to weigh the modalities dynamically. This paper introduces a joint multimodal transformer for emotional recognition. This transformer-based architecture introduces a joint feature representation to add more redundancy and complementary between audio and visual data. The two modalities are first encoded using separate backbones to extract intra-modal spatiotemporal dependencies. The feature vectors of the two modalities are joint and the joint feature vector is also fed into the Joint Multimodal Transformer module. This joint representation provides more fine-grained information about the inter-modal association between the two modalities. The proposed model outperforms state-of-the-art methods on the Biovid dataset and improves over the vanilla multimodal transformer by 6\% on the Affwild2 dataset. Our future work includes introducing more modalities and sophisticated backbones for effective feature extraction.

\section*{Acknowledgement}
This work was supported by the Fonds de recherche du Québec – Santé (FRQS), the Natural Sciences and Engineering Research Council of Canada (NSERC), and Canada Foundation for Innovation (CFI).

\FloatBarrier

\bibliographystyle{apalike}
\bibliography{main}

\end{document}